# An overcome of far-distance limitation on tunnel CCTV-based accident detection in AI deep-learning frameworks

*Kyu-Beom Lee[1] and Hyu-Soung Shin[2]

[1] Smart City & Construction Engineering, UST, Gyeonggi Province 10223, Republic of Korea
[2] Department of Future & Smart Construction Research, KICT, Gyeonggi Province 10223, Republic of Korea
[2] hyushin@kict.re.kr

## ABSTRACT

Tunnel CCTVs are installed to low height and long-distance interval. However, because of the limitation of installation height, severe perspective effect in distance occurs, and it is almost impossible to detect vehicles in far distance from the CCTV in the existing tunnel CCTV-based accident detection system (Pflugfelder 2005). To overcome the limitation, a vehicle object is detected through an object detection algorithm based on an inverse perspective transform by re-setting the region of interest (ROI). It can detect vehicles that are far away from the CCTV. To verify this process, this paper creates each dataset consisting of images and bounding boxes based on the original and warped images of the CCTV at the same time, and then compares performance of the deep learning object detection models trained with the two datasets. As a result, the model that trained the warped image was able to detect vehicle objects more accurately at the position far from the CCTV compared to the model that trained the original image.

## 1. INTRODUCTION

Accidents occurring in tunnels can cause fatal casualties, so it is necessary to quickly detect and respond to accidents in tunnel site. To assist this, some sites are operating a tunnel CCTV-based accident detection system. However, tunnel CCTV inevitably shows severe perspective because it is installed at a low installation height because of the spatial limitation of the tunnel. Also, because of the low illumination in tunnels, the detection performance of the existing computer vision algorithm at a far distance from the tunnel CCTV used to be poor. In fact, Pflugfelder showed that the calculated object size in computer vision theory and the actual measured object size were compared according to the tunnel CCTV and distance, and the error increased significantly as the distance increased (Pflugfelder 2005). To overcome this problem,

---

[1] Graduate Student
[2] Professor & Director

Xu proposed a method to improve the low illumination tunnel image environment by combining stereo vision and image deblurring algorithms (Xu 2017). However, to date, there has been no suggestion to fundamentally reduce the perspective effect in the tunnel CCTV-based accident detection systems. Therefore an attempt in this study is performed to overcome the perspective limitation in tunnels.

## 2. APPLIED METHODOLOGIES

In this paper, to solve the above-mentioned problem, inverse perspective transform that can alleviate the perspective effect is applied to detect vehicle and pedestrian objects on tunnel CCTV. Then, rectangular-shaped bounding boxes are used for labelling and detecting the target objects on warped images.

*2.1 Inverse perspective transform*

An Inverse perspective transform is often adopted in computer vision technology for alleviating a perspective effect by warping images (Quan 1989). It is widely used in autonomous driving and CCTV surveillance area (Liu 2012). In this paper, region of interest (ROI) is required to warp by an inverse perspective transform. ROI is composed of 4 points that are the sum of 2 points on both sides of the edge lane of the road at the nearest distance from the tunnel CCTV and 2 points on both sides of the edge lane of the road at a specified distance. And then the inverse perspective transform is performed with 4 edges and 4 apex points of ROI. And the inverse perspective transform requires a Homography matrix which can be expressed as the following Eq. (1).

$$H = \begin{bmatrix} h_{11} & h_{12} & h_{13} \\ h_{21} & h_{22} & h_{23} \\ h_{31} & h_{32} & 1 \end{bmatrix}$$

(1)

Then, using H as shown in Eq. (2), the position transformation can be performed from the point (x,y) of the original image coordinate system to the point (x',y') of the warped image coordinate system.

$$S \begin{bmatrix} x' \\ y' \\ 1 \end{bmatrix} = H \begin{bmatrix} x \\ y \\ 1 \end{bmatrix} = \begin{bmatrix} h_{11} & h_{12} & h_{13} \\ h_{21} & h_{22} & h_{23} \\ h_{31} & h_{32} & 1 \end{bmatrix} \begin{bmatrix} x \\ y \\ 1 \end{bmatrix}$$

(2)

where S is a scale factor which is used to adjust the scale of the two image coordinate systems to the same value.

*2.2 Deep learning-based object detection model*

In recent years, a deep learning-based object detection model has appeared to guarantee a faster operation speed and far superior performance in comparison with conventional computer vision models. This paper uses Faster R-CNN (Faster Region-based Convolutional Neural Network), the most widely used deep learning-based object detection model, which shows high object detection performance compared to other algorithms and deep learning-based models (Ren 2016).

## 3. COMPARITIVE DEEP-LEARNING EXPERIMENTS

In this paper, by comparing the object detection performance of two deep learning-based object detection models that have trained each features of original and warped images, a perspective effect on the both deep-learning models is investigated in terms of object detection performance. Test datasets not being trained are composed of 527 images which involves nine moving vehicles. The test datasets are used for performance comparison.

*3.1 Experimental condition*
The same number of objects and images were labeled for both original and warped models. Also, since the original images and the warped images should have the same detection range, a ROI was set in the original images as shown in Fig. 1, and then the original image to be compared and the warped image were created as shown in Fig. 1 (a) and (b). Both images are cropped from the original image according to the size of the rectangle consisting of the minimum and maximum points of the ROI. Fig. 1 (a) shows that the predicted image area is allocated by setting the ROI only in the original image as case1 where background out of the ROI is removed in black, as shown in Fig. 1(a). Fig. 1(b) shows a warped image as case2 obtained by the inverse perspective transform process described in chapter 2.1. Since both pictures are cropped images, they are in the same image resolution, and image processing was performed focusing on the image within the ROI. And since detection performance should be investigated in distance, test datasets were composed separately for 4 sections in distance, as shown Fig. 2.

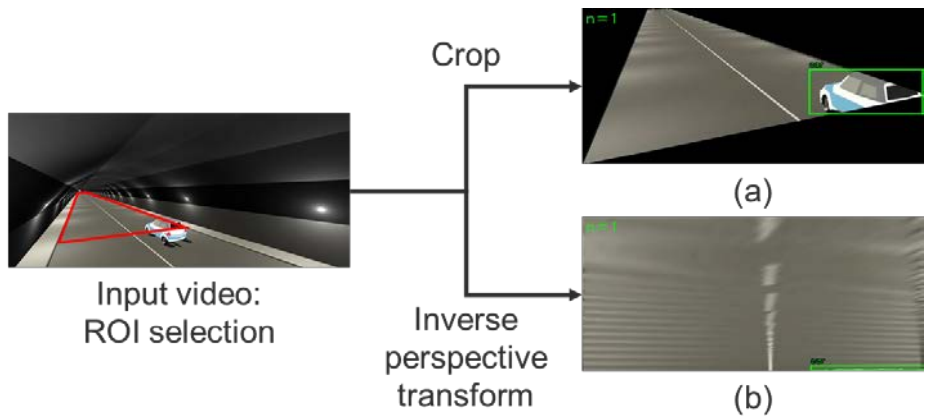

Fig.1 ROI set in the original image by the method proposed in this paper and the image conditions for each case (a) case 1 (b) case 2

Fig. 2 (a) refers to case 1, and Fig. 2 (b) refers to case 2. As shown in Fig. 2, section 1 is the closest section from tunnel CCTV and divided by a line perpendicular to the tunnel driving direction. All the sections are evenly divided by 50m in driving direction. In Fig. 2, x-axis is in direction of cross-section of road and y-axis is in driving direction. Vehicle objects in each section are classified based on the average value of the y-minimum and y-maximum values of the bounding box of target objects. In comparison to Fig. 2(a) and Fig. 2(b), the object size should be the largest in section 1, and the size becomes smaller rapidly as it is moving into section 4. On the other hand, as shown in Fig. 2(b), intervals of sections seem to be similar and image sizes of an moving object in all the sections are relatively uniform.

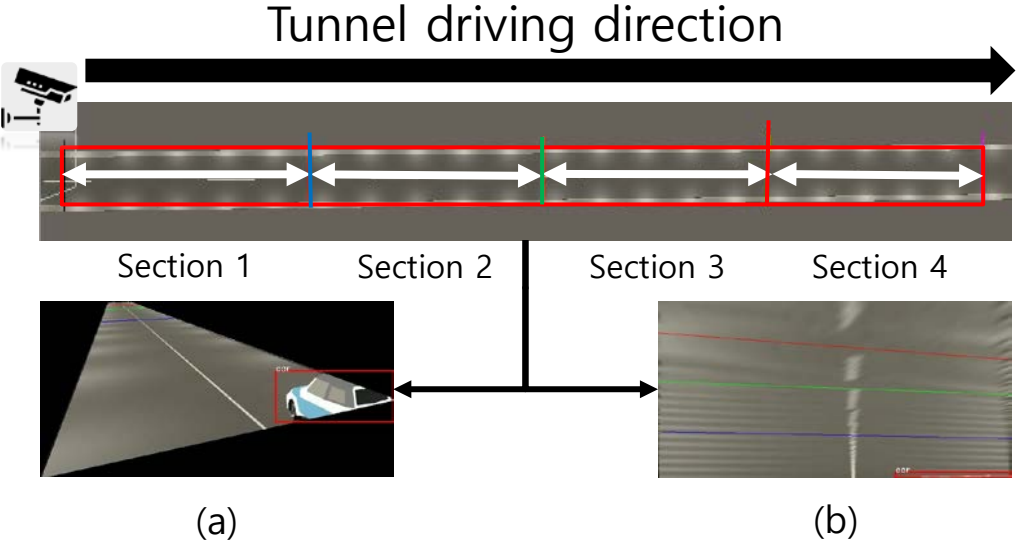

Fig.2 After dividing 2 images of test dataset each 50m interval into 4 sections (a) case 1 (b) case 2

Table 1. Deep learning data set status for each case

| Case | Dataset type | Experiment section | Number of images | Number of car objects |
|---|---|---|---|---|
| case 1 | train | - | **421** | **1387** |
| | test | section 1 | 58 | 111 |
| | | section 2 | 55 | 87 |
| | | section 3 | 63 | 83 |
| | | section 4 | 70 | 92 |
| | | total | **106** | **373** |
| case 2 | train | - | **421** | **1387** |
| | test | section 1 | 59 | 104 |
| | | section 2 | 63 | 91 |
| | | section 3 | 62 | 83 |
| | | section 4 | 67 | 95 |
| | | total | **106** | **373** |

### 3.2 Deep learning dataset

As described in the above chapter, a labeling data set was produced for each case as shown in Table 1. In both cases, the image width and height are 646 and 324 pixels, respectively, and the dataset type is divided into a train dataset to train a deep learning model in the entire dataset and a test dataset to evaluate the final performance of the trained deep learning model. The ratio of the train and test dataset was divided by 8:2. For the deep learning model evaluation, test dataset is separated by each section. For deep learning model training, the learning rate was 0.0001, the batch size for the image was 1, training epoch was 100 and the convolutional layer of the ResNet 50 layer was used. The hardware specifications used for deep learning training are AMD RYZEN 1800X, NVIDIA GTX 1080TI 11GB. The versions of computing environment are Python 3.7 and Tensorflow 1.15.5.

### 3.3 Experimental result

The performance evaluation of the deep learning model is conducted with average precision (AP) (Salton 1983). And the AP value was measured by setting the Intersection over union (IOU) criterion to 0.5, and the test dataset was used for evaluating the performance of the finally completed deep learning models. However, since the deep learning model were trained for all 4 sections, vehicle objects are inferred regardless of the sections. Therefore, the bounding box data deduced outside the range of each section is not used in the calculation of the AP value. Fig. 3 shows comparison of AP values for each section between the 2 cases of deep learning models called original and warped image models after 100 epochs. As shown in Fig. 3, even if the AP value of case 2 is slightly lower than the one of case 1, the detection performances of case 2 are getting better than case 1 as moving into section 4 and also the detection performances in all the sections appear to be relatively consistent.

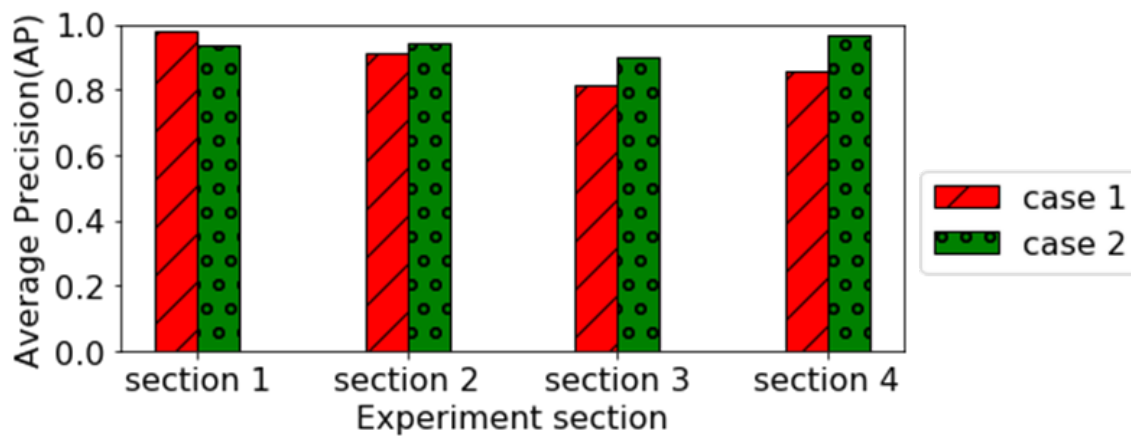

Fig. 3. AP value result according to the inference by section of the deep learning model for each case in the test dataset

### 4. CONCLUSIONS

In this paper, an attempt has been made for overcoming incorrigible perspective limitation on object detection in tunnels. For this, a new concept of inverse perspective transform was adopted for warping original images on tunnel CCTV. This method allows to make all the object sizes similar regardless of distance from CCTV location in tunnel. Then, a comparative study was performed to show the benefit of the use of warping images rather than original image used. Therefore, it seems that detection performance could be consistent even in far distance to 200m from the position of CCTV installation in acceptable detection accuracy (AP) of over 0.85. It means that additional CCTV installation may not be necessary even with operation of automatic accident detection system in tunnels (c.f. 200-250m installation intervals of CCTVs in tunnel in Korean regulation. However, 100m installation interval required for operation of automatic accident detection system).